\pdfoutput=1
\documentclass[11pt]{article}

\usepackage[preprint]{acl}
\usepackage{array}
\usepackage{times}
\usepackage{latexsym}
\usepackage{multirow} 
\usepackage[T1]{fontenc}
\usepackage{hyperref}

\usepackage[utf8]{inputenc}
\usepackage[table]{xcolor}
\definecolor{rowWODiT}{HTML}{EEF4FB}
\definecolor{rowFull}{HTML}{E1E8F0}
\definecolor{rowW4A4}{HTML}{C3D2E6}
\definecolor{rowOurs}{HTML}{DCEAD6}
\usepackage{microtype}
\usepackage{booktabs}
\usepackage{inconsolata}

\usepackage{graphicx}
\usepackage{amssymb}
\usepackage{amsmath}
\newcommand{\methodname}{\texttt{$\Omega$-QVLA}}

\title{\methodname{}: Robust Quantization for Vision-Language-Action Models via Composite Rotation and Per-step Scaling}

\author{
  \textbf{Xinyu Wang\textsuperscript{1,6,*}},
  \textbf{Mingze Li\textsuperscript{2,*}},
  \textbf{Sicheng Lyu\textsuperscript{1,5,6,*}},
  \textbf{Dongxiu Liu\textsuperscript{3}},
  \textbf{Kaicheng Yang\textsuperscript{4}},
\\
  \textbf{Ziyu Zhao\textsuperscript{1}},
  \textbf{Yufei Cui\textsuperscript{1,5}},
  \textbf{Xiao-Wen Chang\textsuperscript{1}},
  \textbf{Peng Lu\textsuperscript{2,\dag}}
\\
\\
  \textsuperscript{1}McGill University,
  \textsuperscript{2}Université de Montréal,
  \textsuperscript{3}Beijing University of Posts and Telecommunications,
\\
  \textsuperscript{4}Shanghai Jiao Tong University,
  \textsuperscript{5}Mila -- Quebec AI Institute,
  \textsuperscript{6}SimpleWay.ai
}

\usepackage{amsmath,amsfonts,bm}

\def\eqref#1{equation~\ref{#1}}

\def\1{\bm{1}}

\def\mR{{\bm{R}}}

\def\mU{{\bm{U}}}

\def\mW{{\bm{W}}}
\def\mX{{\bm{X}}}

\DeclareMathAlphabet{\mathsfit}{\encodingdefault}{\sfdefault}{m}{sl}
\SetMathAlphabet{\mathsfit}{bold}{\encodingdefault}{\sfdefault}{bx}{n}

\begin{document}
\maketitle
\begingroup
\renewcommand{\thefootnote}{\fnsymbol{footnote}}
\footnotetext[1]{
  Equal contribution. Emails:
  \href{mailto:xinyu.wang5@mail.mcgill.ca}{xinyu.wang5@mail.mcgill.ca},
  \href{mailto:mingzeli996@gmail.com}{mingzeli996@gmail.com},
  \href{mailto:sicheng.lyu@mail.mcgill.ca}{sicheng.lyu@mail.mcgill.ca}.
}
\footnotetext[2]{
  Corresponding author. Email:
  \href{mailto:peng.lu@umontreal.ca}{peng.lu@umontreal.ca}.
}
\endgroup
\begin{abstract}
Vision-Language-Action (VLA) models unify perception, reasoning, and control within a single policy, yet their multi-billion-parameter backbones and diffusion-based action heads make on-device deployment prohibitively expensive. Prior quantization efforts offer only partial solutions---compressing the LLM backbone while leaving the DiT action head at full precision, or resorting to mixed-precision schemes---driven by the belief that uniformly quantizing the action head is inherently unstable. We challenge this assumption with \methodname, the first training-free post-training quantization framework that compresses both the language backbone and the entire diffusion action head of a VLA model to a uniform W4A4 precision, eliminating the need for mixed-precision allocation. \methodname{} combines (1) a composite SVD$\cdot$Hadamard rotation that equalizes per-channel weight energy while diffusing residual activation outliers, with (2) per-step DiT activation scaling quantization that absorbs dynamic-range drift across denoising steps. On LIBERO, \methodname{} compresses Pi~0.5 and GR00T~N1.5 to W4A4 with \textbf{98.0\%} and \textbf{87.8\%} task success rates---matching or exceeding their FP16 references (97.1\%, 87.0\%)---while reducing the static memory footprint by \textbf{71.3\%}. Real-world manipulation experiments further confirm smooth, accurate manipulation where prior methods fail. Code is available at \href{https://github.com/UCMP13753/Omega-QVLA}{GitHub}.
\end{abstract}

\section{Introduction}
Vision-Language-Action (VLA) models~\cite{kim2024openvla, intelligence2025pi_, bjorck2025gr00t} unify visual perception, language reasoning, and action generation within a single policy, offering a promising path toward generalist robot control. By inheriting pretrained backbones from large language models (LLMs)~\cite{touvron2023llama} and vision-language models (VLMs)~\cite{li2022blip}, VLA systems can parse natural-language instructions, reason over visual observations, and output executable actions in one forward pass. However, this unification comes at a steep cost: deploying a VLA policy on real robot hardware requires running a multi-billion-parameter language backbone alongside a diffusion-based action head~\cite{chi2025diffusion} under tight latency and memory budgets. Making these models efficient enough for on-device deployment is therefore an urgent and open problem.

Post-training quantization (PTQ)~\cite{xiao2023smoothquant, lin2024duquant} has become the standard recipe for compressing LLMs and VLMs without retraining. Yet a striking gap emerges when we survey existing work on VLA efficiency: \emph{no prior method quantizes the full DiT action head of a VLA model at a uniform low-bit precision.} Across all prior work we examined, the DiT-based action head either remains at full precision or is only selectively quantized---and the authors are explicit about why: they consider it too sensitive to compress. This is no coincidence. Unlike the language backbone, the action head produces continuous control signals that directly interface with physical actuators in a closed-loop setting. Quantization-induced perturbations that would be imperceptible in a language benchmark are amplified by contact forces and physical dynamics. Mainstream PTQ methods---built around outlier management via rotations, permutations, or saliency-based protections---offer no mechanism to account for this cumulative error propagation. 

We challenge the premise that the DiT action head cannot survive uniform low-bit quantization. Rather than routing around the action head's sensitivity, we address it directly at its source: channel-level energy imbalance and denoising-step dynamic-range drift. We present \methodname{}, the first training-free PTQ framework to compress both the language backbone and the entire DiT action head of a VLA model to a uniform W4A4 precision---eliminating the need for mixed-precision allocation altogether. Our contributions are threefold:
\begin{enumerate}
    \item We introduce \methodname{}, a post-training quantization framework combining a composite SVD$\cdot$Hadamard rotation that equalizes per-channel weight energy and diffuses activation outliers, with per-step DiT activation calibration that absorbs dynamic-range drift across denoising steps.

    \item On the LIBERO benchmark~\cite{liu2023libero}, \methodname{} compresses Pi~0.5 and GR00T~N1.5 to W4A4 with average task success rates of \textbf{98.0\%} and \textbf{87.8\%}, matching or exceeding their FP16 references (97.1\% and 87.0\%), while reducing the static memory footprint by \textbf{71.3\%}.

    \item We validate \methodname{} through real-world bimanual manipulation experiments, demonstrating smooth and accurate control under uniform W4A4 quantization where prior methods exhibit degraded performance.
\end{enumerate}
\section{Related Work}

\subsection{LLM quantization}
The core challenge in LLM quantization is activation outliers: a small number of channels carry disproportionately large values that dominate the quantization error. SmoothQuant~\cite{xiao2023smoothquant} handles this by rescaling activations against their corresponding weights, pushing the quantization difficulty from activations to weights, which are easier to compress. Rotation-based methods---QuaRot~\cite{ashkboos2024quarot}, FlatQuant~\cite{sun2024flatquant}, DuQuant~\cite{lin2024duquant}, OSTQuant~\cite{hu2025ostquant}---take a different approach: they apply orthogonal transformations to spread outliers evenly across channels before quantization, so no single dimension suffers excessive error. Weight-only schemes like AWQ~\cite{lin2024awq} and GPTQ~\cite{frantar2022gptq} sidestep activation quantization entirely by compressing only the weights, which works well for memory-bound deployment but leaves compute-bound scenarios unaddressed.

\subsection{DiT quantization}
Unlike standard autoregressive language models, diffusion transformers (DiTs) exhibit highly dynamic activation statistics that fluctuate continuously across the multi-step denoising process. Recognizing this severe limitation, recent research has pivoted toward developing bespoke, DiT-specific compression strategies. For example, SVDQuant~\citep{li2024svdquant} addresses this temporal variance by absorbing prominent activation outliers through the addition of full-precision, low-rank branches. Similarly, ViDiT-Q~\citep{zhao2024vidit} introduces a granular, per-group dynamic quantization method that adaptively captures and adjusts to the shifting activation ranges at each specific timestep. The fundamental inadequacy of directly transferring text-based techniques is systematically formalized by the PTQ4DiT benchmark~\citep{wu2024ptq4dit}. Their extensive evaluations definitively show that off-the-shelf LLM quantization recipes degrade both the spatial and temporal consistencies of diffusion models.

\subsection{VLA quantization}
Progress in one component of a VLA system does not automatically transfer to others. For example, improvements in quantization techniques for LLMs do not guarantee similar gains in multimodal or action‑generation modules, and methods designed for one modality often fail to ensure efficient or stable behavior when applied to the full VLA pipeline. QuantVLA~\citep{quantvla}, QVLA~\citep{xu2026qvla}, and similar recent efforts address this gap by developing compression strategies explicitly tailored for the unique architectural and operational demands of embodied control. Specifically, QuantVLA introduces a scale-calibrated post-training quantization framework that selectively quantizes both the language backbone and the diffusion transformer (DiT) action head to preserve operator schedules and control fidelity. Taking a complementary approach, QVLA highlights the compounding execution errors caused by naive, uniform-bit LLM quantization, proposing instead an action-centric, channel-wise bit allocation method that explicitly measures and minimizes action-space sensitivity. Collectively, these works demonstrate that effectively compressing VLA systems requires holistic, action-aware optimization rather than the direct transfer of language-centric techniques.

\begin{figure*}[t!]
    \centering
    \includegraphics[width=0.85\linewidth]{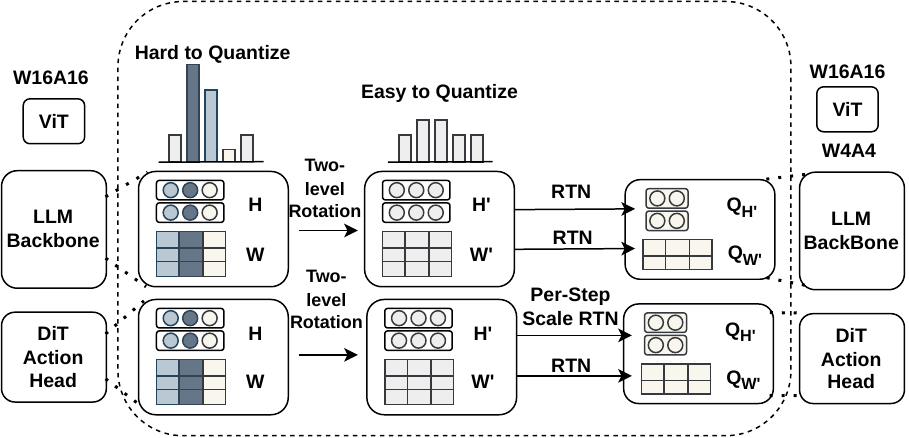}
    \caption{The overall quantization pipeline of our \methodname{}, where RTN means round-to-nearest-neighbor.}
    \label{fig:aspq}
\end{figure*}
\section{Background}
\subsection{Vision Language Action (VLA) Model}
VLA models form a class of embodied systems that jointly process visual observations, interpret textual instructions, and generate action representations executable on physical robotic platforms operating in dynamic environments. Similar to models that map sensory inputs and prompts to action sequences, VLAs integrate vision encoders (e.g., ViTs), language models (e.g., LLMs), and action-generation modules or planners to produce task-conditioned behaviors (e.g., DiTs). Their architectures typically extend multimodal fusion techniques established in vision--language models—such as cross-attention, concatenated embeddings, or token-level unification—to align sensory inputs, linguistic directives, and action representations within a unified control framework.

\subsection{Model Quantization}
Quantization maps full‑precision parameters $\mW$ to a low‑precision representation $\hat{\mW}=Q(\mW)$ through a discrete transformation. The objective is to design a quantizer that preserves model behavior under reduced precision. Consider a weight matrix $\mathbf{W}$ from a linear layer $\mathbf{Y} = \mathbf{XW} + \mathbf{b}$. Uniform quantization computes a single scaling factor $\alpha_W$ for the entire tensor and maps all full‑precision values to $k_w$‑bit integers $\mW_q$.
The de‑quantized weights are then recovered as $\hat{\mathbf{W}} = \alpha_W \mathbf{W}_q$, yielding an approximation of the original computation, \emph{i.e.}, $\mathbf{Y} \approx \hat{\mathbf{X}}\hat{\mathbf{W}} + \mathbf{b}$, where $\hat{\mathbf{X}}$ denotes the quantized activations.
To accelerate computation, modern commercial GPUs require both weights and activations to be quantized to the same bit width. Otherwise, if the bit widths differ, the lower-precision values must be upcast to match the higher-precision ones (or vice versa), which negates the performance benefits of quantization.

\section{Methodology}
\label{sec:method}

In this work, we propose \methodname, a post-training quantization framework for VLA models with a unified two-stage pipeline for the LLM backbone and DiT action head. \methodname{} mitigates this via: (1) \textit{Two-level Rotation}, using a composite SVD$\cdot$Hadamard rotation to suppress channel-level outliers before quantization; and (2) Per-step DiT Activation Scaling Quantization to handle timestep-dependent shifts from iterative denoising.

\subsection{VLA Quantization via Composite Rotation}
\label{sec:composite_rotation} 
\begin{figure*}[t!]
  \includegraphics[width=1\linewidth]{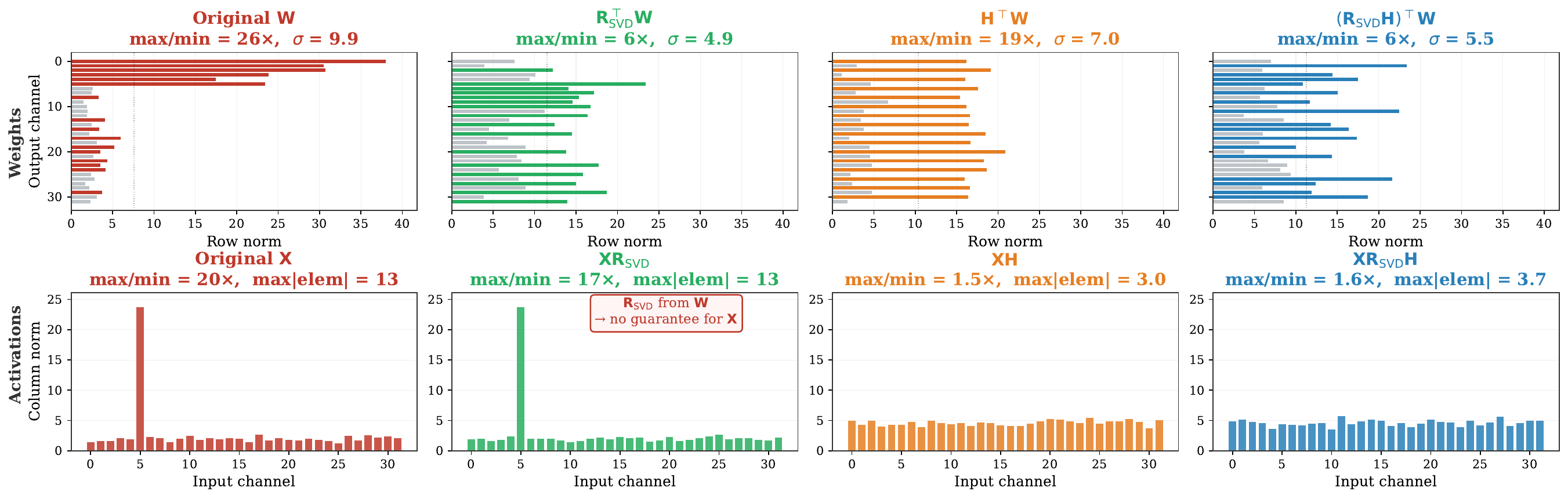}
              \caption{Per-channel distribution of weights (top) and activations (bottom) under four rotation settings. $\mathbf{R}_{\mathrm{SVD}}$, adapted to $\mathbf{W}$, equalizes weight row norms effectively (26$\times$ $\to$ 6$\times$, $\sigma$: 9.9 $\to$ 4.9), whereas the data-independent $\mathbf{H}$ provides limited improvement (19$\times$, $\sigma$=7.0). Conversely, $\mathbf{H}$ smooths activations (20$\times$ $\to$ 1.5$\times$) via $\lVert \mathbf{z}\mathbf{H} \rVert_\infty \leq \lVert \mathbf{z} \rVert_2 / \sqrt{n}$, while $\mathbf{R}_{\mathrm{SVD}}$, derived entirely from $\mathbf{W}$, leaves activation outliers intact (20$\times$ $\to$ 17$\times$). The composite $\mathbf{R}_{\mathrm{SVD}}\mathbf{H}$ inherits both strengths (weights: 6$\times$, $\sigma$=5.5; activations: 1.6$\times$). Note that randomized Hadamard $\mathbf{H}\mathbf{D}$ ($\mathbf{D}=\mathrm{diag}(\pm1)$) yields identical per-channel norms, since sign flips do not alter row or column norms.}

    \label{fig:composite_rotation}
\end{figure*}
Rotation-based quantization mitigates outliers by applying an orthogonal transformation before quantization. Given a linear layer $\mathbf{Y}=\mathbf{X}\mathbf{W}$, where $\mX$ is the input and $\mW$ is the weight matrix, it can be rewritten as $\mathbf{Y}=(\mathbf{X}\mathbf{R})(\mathbf{R}^{\top}\mathbf{W})$ with $\mathbf{R}\mathbf{R}^{\top}=\mathbf{I}$, enabling quantization in a rotated space with better statistical properties. In per-channel low-bit quantization, large inter-channel variance causes high-energy channels to be quantized too coarsely while wasting precision on low-energy channels. Reducing this variance improves overall bit utilization. We focus on W4A4 quantization, where both weight and activation outliers are especially challenging. Therefore, we propose a two-level rotation method to address the outliers for weight and activation, respectively.

\paragraph{SVD rotation.}
Given a weight matrix $\mathbf{W} \in \mathbb{R}^{C_{\text{in}} \times C_{\text{out}}}$, we compute its singular value decomposition:
\begin{align}
    \mathbf{W} &= \mathbf{U}\boldsymbol{\Sigma}\mathbf{V}^{\top}, \\
    \mathbf{R}_{\mathrm{SVD}} &= \mathbf{U}.
\end{align}
Applying the rotation yields
$
\widetilde{\mathbf{W}}
=
\mathbf{R}_{\mathrm{SVD}}^{\top}\mathbf{W}
=
\mathbf{U}^{\top}\mathbf{W}
=
\boldsymbol{\Sigma}\mathbf{V}^{\top}$. In the original parameterization, the row-wise energy of $\mathbf{W}$ is $\|\mathbf{w}_i\|_2^2
=\sum_{k=1}^{r}\sigma_k^2 u_{ik}^2$, where $u_{ik}$ denotes the $(i,k)$-th entry of $\mathbf{U}$. Thus, when the left singular vectors are unevenly aligned with the coordinate basis, dominant singular values may concentrate disproportionately in a few rows, producing channel-wise outliers and large inter-channel variance. After SVD rotation, the rotated weight matrix satisfies
$
\|\widetilde{\mathbf{w}}_i\|_2^2
=
\sigma_i^2$, meaning that each row energy is directly controlled by the singular spectrum rather than coordinate-dependent mixing. Consequently, the rotation removes coordinate-induced energy concentration, yielding a smoother row-wise magnitude distribution that is more amenable to low-bit quantization (Figure~\ref{fig:composite_rotation}, top row).

\paragraph{Hadamard rotation.}
Since $\mathbf{R}_{\mathrm{SVD}}$ is derived entirely from $\mathbf{W}$, it provides no guarantee on the activation side: whether $\mathbf{X}\mathbf{U}$ has a more uniform channel distribution than $\mathbf{X}$ depends on the alignment between the activation outliers and $\mathbf{U}$'s column structure, which is data-dependent and can leave activation outliers intact (Figure~\ref{fig:composite_rotation}, bottom middle). To mitigate this issue, we therefore compose $\mathbf{R}_{\mathrm{SVD}}$ with a normalized Hadamard matrix $\mathbf{H} \in \{\pm 1/\sqrt{C_{\text{in}}}\}^{C_{\text{in}} \times C_{\text{in}}}$, defining the composite rotation as: \begin{equation}
    \mathbf{R} \;=\; \mathbf{R}_{\mathrm{SVD}} \cdot \mathbf{H}.
    \label{eq:composite_rot}
\end{equation}
As a uniform mixing matrix, $\mathbf{H}$ redistributes the energy of any dominant channel uniformly across all $n$ output channels. Formally, for any $z \in \mathbb{R}^n$:

\paragraph{Block-wise Implementation.}
Nevertheless, directly instantiating $\mathbf{R} = \mathbf{R}_{\mathrm{SVD}} \cdot \mathbf{H}$
as a dense $C_{\text{in}} \times C_{\text{in}}$ matrix is computationally
prohibitive and incurs substantial memory overhead.
We instead approximate $\mathbf{R}$
in a block-wise manner:
\begin{align}
    \hat{\mathbf{R}} \;=&\; \operatorname{BlockDiag}(\mR_1, \dots, \mR_K),
    \\
    \mR_b = & \;  \mU_{\mathrm{SVD}}\!\left(\mathbf{W}_b^\top\right) \cdot \mathbf{H}_c,
    \label{eq:blockdiag}
\end{align}
where each $\mathbf{W}_b$ is a contiguous row partition of $\mathbf{W}$, $c$ is the corresponding row dimension of $\mW_b$, $\mU_{\text{SVD}}$ is the matrix of left singular vectors.
However, since each block rotation operates independently within its partition, outlier channels clustered within the same block remain jointly suppressed under a single group scale, thus the magnitude of different blocks could have large difference. To avoid outliers concentrate on some blocks,  we follow ~\citet{lin2024duquant} and introduce an orthogonal permutation matrix $\mathbf{P}$ satisfying $\mathbf{P}\mathbf{P}^\top = \mathbf{I}$ that reorders channels prior to block partitioning. We sort channels by descending weight column norm $\lVert \mathbf{W}_{j,:} \rVert_2^2$ and distribute them across the $K$ blocks in a zigzag order, such that each block receives channels of alternating high and low norms rather than a homogeneous range. This balances the energy distribution across blocks, preventing high-norm channels from concentrating within any single partition. The complete transformation is then:
\begin{align}
    \mathbf{X}' = &\mathbf{X}\,\mathbf{P}\,\hat{\mathbf{R}}, \\
    \mathbf{W}' = &\mathbf{P}^\top\hat{\mathbf{R}}^\top\,\mathbf{W},
    \label{eq:transform}
\end{align}
where $\mathbf{P}$ is determined solely by weight column norms, without requiring activation statistics, to avoid overfitting to the calibration
distribution.

\subsection{Per-step DiT Activation Scaling Quantization}
\label{sec:quant_solver}

Quantization is applied after the composite rotation, yielding the
rotated weight $\mathbf{W}'$ and activation $\mathbf{X}'$. Both are
quantized using symmetric uniform quantization. For a generic
tensor $\mathbf{Z}$ at bit-width $k$, let
$q_{\max}=2^{k-1}-1$ and define the scale and quantized
representation as:
\begin{align}
    \Delta_{\mathbf{Z}}
    &= \frac{\max(|\mathbf{Z}|)}{q_{\max}},
    \label{eq:scale} \\
    \mathbf{Q}_{\mathbf{Z}}
    &= \mathrm{clamp}\!\Big(
        \!\left\lfloor
            \tfrac{\mathbf{Z}}{\Delta_{\mathbf{Z}}}
        \right\rceil,\,
        -q_{\max},\,q_{\max}
    \Big).
    \label{eq:quant}
\end{align}
The dequantized value is reconstructed as
$Q(\mathbf{Z})=\Delta_{\mathbf{Z}}\,\mathbf{Q}_{\mathbf{Z}}$.
For weights, $\Delta_{\mathbf{W}'}$ is computed per output
channel at bit-width $k_w$; for LLM backbone activations,
$\Delta_{\mathbf{X}'}$ is computed per token at bit-width $k_a$.

However, a single static scale is suboptimal for the DiT action
head, whose activation magnitudes vary substantially across
denoising steps. We therefore introduce \emph{per-step activation
calibration}: a per-step, per-layer, per-channel scale table
precomputed offline from calibration trajectories. For layer
$\ell$, step $t$, and channel $j$:
\begin{align}
    \Delta_{\ell,t,j}
    = \frac{\hat{\sigma}\!\big(
        \mathbf{X}'^{(\ell)}_{t,:,j}
    \big)}{q_{\max}},
    \label{eq:perstep_scale}
\end{align}
where $\hat{\sigma}(\cdot)$ is a robust peak estimator. At
inference, the scale for the current step is retrieved and the
activation is quantized via Eq.~\ref{eq:quant} with
$\Delta_{\ell,t,\cdot}$ in place of $\Delta_{\mathbf{Z}}$.
The low-bit linear operation is then:
\begin{align}
    \mathbf{X}'\mathbf{W}'
    \approx
    \Delta_{\mathbf{X}'}
    \Delta_{\mathbf{W}'}
    \,\mathbf{Q}_{\mathbf{X}'}
    \mathbf{Q}_{\mathbf{W}'}.
    \label{eq:lowbit_linear}
\end{align}

\section{Experiments and Results}
In this section, we describe our experimental setup and report
results on both simulation benchmarks and real-world manipulation
tasks, evaluated across two VLA model families.
\begin{table*}[t!]
\centering
\fontsize{5.3pt}{6.1pt}\selectfont
\setlength{\tabcolsep}{2.0pt}
\renewcommand{\arraystretch}{0.98}
\vspace{0.2em}
\resizebox{\textwidth}{!}{
\begin{tabular}{@{}l c l ccccc!{\vrule width 0.5pt}ccccc@{}}
\toprule
\multirow{2}{*}{\textbf{Method}} 
& \multirow{2}{*}{\textbf{Precision}} 
& \multirow{2}{*}{\textbf{Quantization Setting}}
& \multicolumn{5}{c}{\textbf{GR00T N1.5}}
& \multicolumn{5}{c}{\textbf{Pi 0.5}} \\
\cmidrule(lr){4-8}\cmidrule(lr){9-13}
& &
& \textbf{Goal} & \textbf{Spatial} & \textbf{Object} & \textbf{Long} & \textbf{Average}
& \textbf{Goal} & \textbf{Spatial} & \textbf{Object} & \textbf{Long} & \textbf{Average} \\
\midrule

Full Precision & FP16 & No quantization
& 86.0 & 92.0 & 92.0 & 76.0 & 87.0
& 98.5 & 99.0 & 97.5 & 93.5 & 97.1 \\

\midrule
\rowcolor{rowWODiT}
GPTQ & W4A16 & w/o DiT attention
& 81.0 & 91.0 & 81.0 & 68.0 & 80.3
& 76.0 & 85.0 & 91.0 & 75.0 & 81.8 \\
\rowcolor{rowWODiT}
AWQ & W4A16 & w/o DiT attention
& 51.0 & 92.0 & 95.0 & 75.0 & 78.3
& 67.0 & 85.0 & 92.0 & 82.0 & 81.5 \\
\rowcolor{rowWODiT}
OmniQuant & W4A16 & w/o DiT attention
& 86.0 & 90.0 & 68.0 & 76.0 & 80.0
& 74.0 & 85.0 & 94.0 & 78.0 & 82.8 \\

\cmidrule(lr){1-13}
\rowcolor{rowFull}
GPTQ & W4A16 & Full quantization
& 49.0 & 43.0 & 73.0 & 60.0 & 56.3
& 10.0 & 15.0 & 39.0 & 0.0 & 16.0 \\
\rowcolor{rowFull}
AWQ & W4A16 & Full quantization
& 51.0 & 38.0 & 74.0 & 64.0 & 56.8
& 11.0 & 4.0 & 29.0 & 2.0 & 11.5 \\
\rowcolor{rowFull}
OmniQuant & W4A16 & Full quantization
& 43.0 & 39.0 & 65.0 & 53.0 & 50.0
& 10.0 & 8.0 & 23.0 & 0.0 & 10.3 \\

\cmidrule(lr){1-13}
\rowcolor{rowFull}
SmoothQuant & W4A8 & Full quantization
& 89.0 & 82.0 & 94.0 & 80.0 & 86.3
& 96.0 & 99.0 & 98.0 & 94.0 & 96.8 \\
\rowcolor{rowFull}
DuQuant & W4A8 & Full quantization
& 83.0 & 75.0 & 86.0 & 77.0 & 80.3
& 93.0 & 96.0 & 99.0 & 92.0 & 95.0 \\
\rowcolor{rowFull}
QuantVLA & W4A8 & Full quantization
& 53.0 & 44.0 & 50.0 & 77.0 & 56.0
& 90.0 & 92.0 & 98.0 & 56.0 & 84.0 \\
\rowcolor{rowWODiT}
QuantVLA & W4A8 & w/o DiT attention
& 90.0 & 96.0 & 92.0 & 74.0 & 88.0
& 98.0 & 98.5 & 98.0 & 96.0 & 97.6 \\

\cmidrule(lr){1-13}
\rowcolor{rowW4A4}
SmoothQuant & W4A4 & Full quantization
& 83.0 & 87.0 & 84.0 & 82.0 & 84.0
& 40.0 & 83.0 & 88.0 & 26.0 & 59.3 \\
\rowcolor{rowW4A4}
DuQuant & W4A4 & Full quantization
& 62.0 & 66.0 & 74.0 & 78.0 & 70.0
& 94.0 & 96.0 & 99.0 & 88.0 & 94.3 \\
\rowcolor{rowW4A4}
QuantVLA & W4A4 & Full quantization
& 63.0 & 71.0 & 71.0 & 74.0 & 69.8
& 80.0 & 94.0 & 98.0 & 56.0 & 82.0 \\
\rowcolor{rowW4A4}
QuantVLA & W4A4 & w/o DiT attention
& 60.0 & 62.0 & 74.0 & 70.0 & 66.5
& 98.0 & 96.0 & 100.0 & 86.0 & 95.0 \\

\midrule
\rowcolor{rowOurs}
\methodname{} & \textbf{W4A4} & \textbf{Full quantization}
& \textbf{91.0} & \textbf{86.0} & \textbf{92.0} & \textbf{82.0} & \textbf{87.8}
& \textbf{100.0} & \textbf{99.0} & \textbf{97.0} & \textbf{96.0} & \textbf{98.0} \\

\bottomrule
\end{tabular}
}
\caption{
Quantization performance comparison on LIBERO with calibration sample size $n=10$.
All numbers are success rates (\%). ``w/o DiT attention'' denotes excluding DiT attention layers from quantization.
\methodname{} is highlighted in green. 
}
\label{tab:ns10_quant_results}
\end{table*}
\begin{table}[!t]
\centering
\scriptsize
\setlength{\tabcolsep}{3pt}
\vspace{0.3em}
\resizebox{\columnwidth}{!}{
\begin{tabular}{lcccccc}
\toprule
\textbf{Model} 
& \textbf{Cup} 
& \textbf{Blocks} 
& \textbf{Fruit} 
& \textbf{Flowers} 
& \textbf{Towel} 
& \textbf{Avg.} \\
\midrule
Pi-0.5 Base 
& 88.0 & 57.5 & 55.0 & 22.5 & 25.0 & 49.6 \\
QuantVLA 
& 26.0 & 45.0 & 17.0 & 21.0 & 16.0 & 25.0 \\
\textbf{\methodname{}}
& \textbf{84.0} & \textbf{67.5} & \textbf{42.0} 
& \textbf{38.5} & \textbf{23.0} & \textbf{51.0} \\
\bottomrule
\end{tabular}
}
\caption{
Real-world manipulation results using Pi-0.5 under W4A4 quantization.
}
\label{tab:quant_results}
\end{table}
\subsection{Experimental Settings}

\paragraph{Simulation Benchmark.}
We evaluate \methodname{} on two state-of-the-art VLA policies, OpenPI $\pi_{0.5}$~\citep{intelligence2025pi_} and GR00T N1.5~\citep{bjorck2025gr00t}, both employing a DiT-based action head that maps fused visual--language features to action sequences. The two backbones span complementary regimes---$\pi_{0.5}$ prioritizes efficient inference, while GR00T N1.5 offers higher capacity and richer action modeling---enabling a robust assessment across different coupling strengths between perception and control. Evaluation uses the LIBERO~\citep{liu2023libero} simulator with four task suites that target distinct capabilities: \textit{Spatial} tests relational reasoning and precise placement, \textit{Object} focuses on object-centric grasping and manipulation, \textit{Goal} measures instruction-to-goal alignment, and \textit{Long} examines temporal decomposition and control of accumulated error. We follow the standard LIBERO protocol and report success rates on each suite as well as the four-suite average.

\paragraph{Real-World Experiments.}
To verify the effectiveness of \methodname{} on physical hardware, we evaluate on five bimanual manipulation tasks of increasing difficulty: the relatively simple \textit{Pick Cup}, the intermediate \textit{Put Blocks} and \textit{Put Fruit}, and the more challenging long-horizon \textit{Put Flowers} and \textit{Fold Towel}. The experimental setup and task illustrations are shown in Fig.~\ref{fig:real-task-setting}. All experiments are conducted on a dual-arm ARX R5 platform equipped with two 6-DoF arms with parallel-jaw grippers and three RGB camera views. We use Pi-0.5 as the full-precision base model and compare two W4A4 quantization approaches: QuantVLA as the baseline and the proposed \methodname{}. Since each task consists of multiple sequential stages, we adopt a stage-wise progress score for fine-grained evaluation; detailed scoring criteria are provided in Appendix~\ref{sec:real_tasks}. We report the progress score averaged over 10 rollouts per task.

\paragraph{Implementation Details.}
Unless otherwise stated, \methodname{} is applied with a uniform W4A4 precision on all linear layers across language backbone and DiT action head. Following the composite rotation in Section~\ref{sec:method}, we use block-wise SVD$\cdot$Hadamard rotation with block size $64$ and a zigzag weight-norm channel permutation. The LLM side uses GPTQ-style weight quantization (block size $128$, damping ratio $0.01$); the DiT side uses RTN with the per-step activation scale table calibrated over $T=8$ Euler denoising steps. All scales are estimated from a small unlabeled calibration buffer of $n=10$ trajectories and folded into dequantization at inference time. All experiments are run on NVIDIA A100 GPUs. Hyperparameters for \methodname{} and all baselines are detailed in Appendix~\ref{sec:key_hyperparameters}.
\subsection{Main Results}
Table~\ref{tab:ns10_quant_results} reports the simulation results on LIBERO
under calibration sample size $n=10$. We evaluate two representative VLA
backbones, GR00T N1.5 and Pi-0.5, and compare \methodname{} with both
weight-only and weight-activation PTQ baselines. The results show that
full-stack quantization is substantially harder than selectively excluding
DiT attention. Under W4A16, excluding DiT attention keeps the average success
rate around 78--80\% on GR00T N1.5 and 81--83\% on Pi-0.5, whereas quantizing
the full stack causes a sharp drop, especially on Pi-0.5 where GPTQ, AWQ, and
OmniQuant fall to 16.0\%, 11.5\%, and 10.3\%, respectively. This confirms that
DiT attention is a major source of quantization sensitivity in VLA models.

Despite this difficulty, \methodname{} remains effective under the aggressive
W4A4 full-quantization setting. On GR00T N1.5, \methodname{} achieves 87.8\%
average success rate, matching the FP16 reference performance of 87.0\% and
outperforming W4A4 full-quantization baselines such as SmoothQuant, DuQuant,
and QuantVLA. On Pi-0.5, \methodname{} achieves 98.0\% average success rate,
slightly exceeding the FP16 reference of 97.1\%. Compared with QuantVLA under
W4A4 full quantization, \methodname{} improves the average success rate from
82.0\% to 98.0\%, with particularly large gains on the long-horizon suite
(56.0\% $\rightarrow$ 96.0\%). These results indicate that \methodname{}
can quantize the full VLA stack, including the DiT attention layers, without
sacrificing task performance.

Overall, the simulation results demonstrate two key points. First, naively
applying standard PTQ methods to the complete VLA model is unstable, especially
when the action head attention layers are included. Second, the proposed
rotation and per-step calibration design substantially mitigates this failure
mode, allowing W4A4 full-stack quantization to reach or even surpass
full-precision performance on both evaluated backbones.

\begin{figure*}[htbp]
    \centering
    \includegraphics[width=0.95\textwidth]{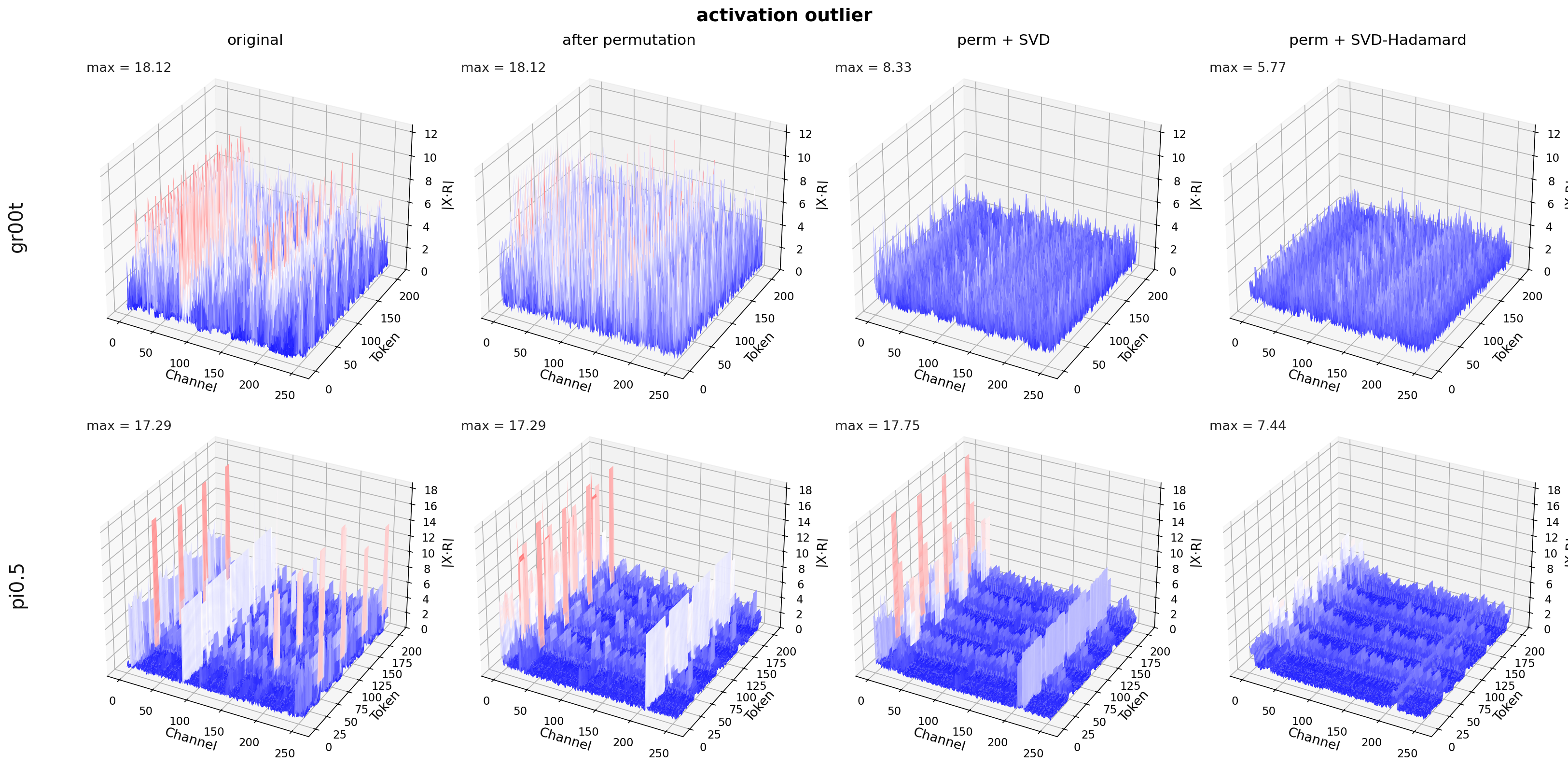}
    \caption{Activation outlier suppression of rotation with SVD~\citep{quantvla} versus our SVD-Hadamard. Per-channel/per-token magnitude surfaces $|\mX\cdot \mR|$ of one representative layer in GR00T-N1.5 (top) and pi0.5 (bottom), shown after applying progressively richer rotations: identity, channel permutation, DuQuant with SVD, and our SVD-Hadamard (which composes a within-block Hadamard transform on top of the SVD basis). The peak magnitude (annotated on each panel) drops monotonically across the pipeline on both models, illustrating how the additional Hadamard step diffuses the residual outliers that survive the SVD rotation — the source of the dynamic-range headroom needed for 4-bit activation quantization.}
    \label{fig:outlier_flow}
\end{figure*}

\subsection{Real-world experiment results}
Table~\ref{tab:quant_results} reports results on five real-world manipulation tasks (task descriptions and scoring criteria in Appendix~\ref{sec:real_tasks}). Under W4A4, \methodname{} reaches an average progress score of 51.0, slightly surpassing the FP16 Pi-0.5 baseline (49.6), while QuantVLA collapses to 25.0 with unreliable execution across multiple tasks. The two policies also differ qualitatively: QuantVLA produces jerky end-effector trajectories that accumulate over long horizons into task failures, whereas \methodname{} yields substantially smoother actions. This behavioral gap is corroborated by the open-loop action analysis in Appendix~\ref{sec:action_analysis}, where \methodname{} tracks the reference trajectories closely while QuantVLA produces frequent spikes across action dimensions.
\section{Discussion and Analysis}
In this section, we present a comprehensive analysis comprising
efficiency benchmarks, ablation studies of individual components,
and an evaluation of the rotation method's effect on quantization
quality.
\subsection{Memory Efficiency and Metadata Overhead}
Table~\ref{tab:memory_savings} reports the static model footprint across quantization configurations. The W4A16 GPTQ baseline establishes the upper bound for memory reduction ($\sim$74\% on both models), since it only stores per-group scaling factors. \methodname{} (W4A4) incurs a negligible overhead of $\sim$90~MB on Pi-0.5 and 56~MB on GR00T-N1.5---accounting for the block-wise SVD rotation matrices, zigzag permutation indices, and per-step activation scale tables---while retaining 72.0\% and 71.3\% savings.

The gap to selective baselines widens on attention-heavy models. QuantVLA quantizes only MLP layers and keeps attention in FP16, so on GR00T-N1.5---where DiT attention occupies a prominent portion of the network---its savings drop sharply to 61.3\%, a 10-point gap below \methodname{}. The gap narrows on Pi-0.5 due to its more compact attention sub-structures, confirming that selective quantization scales poorly with attention allocation. Note that static footprint is dictated entirely by the weight bit-width (W4); A8$\to$A4 affects runtime activation memory rather than disk size.

\begin{table}[t]
\centering
\resizebox{\columnwidth}{!}{%
\begin{tabular}{lccc}
\toprule
\textbf{Configuration} & \textbf{\# Q-Layers} & \textbf{Footprint} & \textbf{Savings} \\ 
\midrule
\multicolumn{4}{c}{\textit{Pi-0.5 (FP16 baseline: 4.27 GB)}} \\
FP16 baseline & 252 & 4.27 GB & 0.0\% \\
W4A16 GPTQ~\citep{frantar2022gptq} & 252 & 1.11 GB & \textbf{74.0\%} \\
QuantVLA W4A8~\citep{quantvla} & 180 & 1.28 GB & 70.0\% \\
\methodname{} & 252 & 1.20 GB & 72.0\% \\ 
\midrule
\multicolumn{4}{c}{\textit{GR00T-N1.5 (FP16 baseline: 1.99 GB)}} \\
FP16 baseline & 180 & 1.99 GB & 0.0\% \\
W4A16 GPTQ~\citep{frantar2022gptq} & 180 & 530 MB & \textbf{74.0\%} \\
QuantVLA W4A8~\citep{quantvla} & 116 & 790 MB & 61.3\% \\
\methodname{} & 180 & 586 MB & 71.3\% \\ 
\bottomrule
\end{tabular}
}
\caption{Static memory footprint and storage savings across quantization configurations. ``\# Q-Layers'' denotes the number of layers executing in the quantized format.}
\label{tab:memory_savings}
\end{table}

\begin{table}[ht]
\centering
\resizebox{\linewidth}{!}{
\begin{tabular}{llcccccc}
\toprule
\textbf{Rotation Matrix} & \textbf{PS} & \textbf{Object} & \textbf{Spatial} & \textbf{Goal} & \textbf{Long} & \textbf{4-avg} \\
\midrule
SVD & \checkmark & 87 & 82 & 73 & 75 & 79.25 \\
SVD$\cdot$Hadamard & $\times$ & \textbf{93} & 83 & \textbf{92} & 75 & 85.75 \\
SVD$\cdot$Hadamard & \checkmark & 91 & \textbf{86} & \textbf{92} & \textbf{82} & \textbf{87.75} \\
\bottomrule
\end{tabular}
}
\caption{Ablation study on the rotation matrix and per-step scaling strategy under the W4A4 setting. All variants use GPTQ for the LLM and RTN for the DiT. "PS" means applying per-step scaling on DiT steps.}
\label{tab:ablation_combined}
\end{table}

\subsection{Ablation Study} 
Table~\ref{tab:ablation_combined} isolates the contributions of our proposed SVD$\cdot$Hadamard rotation and per-step scaling under the W4A4 setting (using GPTQ for the LLM and RTN for the DiT). Replacing the baseline SVD rotation with our combined SVD$\cdot$Hadamard matrix improves the 4-task average by 8.5 points (79.25 $\rightarrow$ 87.75), driven by a remarkable 19-point surge in the \textit{Goal} task. Additionally, disabling per-step scaling degrades the overall average by 2.0 points and triggers a severe 7-point drop on the \textit{Long} suite, as extended horizons strongly amplify inter-step activation drift.

The 8.5-point gain has a clean microscopic explanation: across 10 sampled W4A4 layers, SVD$\cdot$Hadamard achieves the lowest nMSE on 9/10, with $2\text{-}5\times$ reductions over SVD-only (e.g., \texttt{LLM.L02.q\_proj}: $0.139\to 0.028$). The Hadamard mixing diffuses residual single-channel concentration across the 64 channels of each block---the failure mode an SVD basis fitted on weights alone cannot address. One outlier case (\texttt{LLM.L02.down\_proj}, channel-skew $\sim$$5\times10^{4}$) further motivates pairing the rotation with GPTQ on the LLM, while the DiT side uses plain RTN; full layer breakdowns are in Appendix~\ref{app:hadamard} and \ref{app:per_step_scale}.

This design is further validated by a cross-model pattern: SmoothQuant and DuQuant flip orderings between backbones (GR00T: 84.0 vs.\ 70.0; Pi-0.5: 59.3 vs.\ 94.3), reflecting that GR00T's sharply-peaked outliers favor per-channel migration while Pi-0.5's diffuse outliers favor rotation. \methodname{}'s block-wise SVD$\cdot$Hadamard handles both regimes uniformly---SVD adapts to each block's local outlier basis, Hadamard spreads residual energy across 64 channels---and is the only entry in Table~\ref{tab:ns10_quant_results} that simultaneously reaches 87.8\% on GR00T and 98.0\% on Pi-0.5.

\subsection{Smoothing Effect of Composite Rotation}
Figure~\ref{fig:outlier_flow} illustrates activation outlier suppression via rotation with SVD~\citep{quantvla} compared to the proposed SVD-Hadamard method. The figure presents per-channel and per-token magnitude surfaces $|\mathbf{X} \cdot \mathbf{R}|$ for a representative layer in the GR00T-N1.5 (top row) and pi0.5 (bottom row) models. The surfaces are shown after the application of progressively more expressive rotations: identity, channel permutation, DuQuant with SVD, and the proposed SVD-Hadamard transform, which composes a within-block Hadamard transform with the SVD basis. The peak magnitude, annotated in each panel, decreases monotonically across the sequence of transformations for both architectures. This consistent reduction demonstrates that the additional Hadamard step effectively diffuses residual outliers that persist after the SVD rotation, thereby providing the dynamic-range headroom necessary for 4-bit activation quantization.
\section{Conclusion}
We presented \methodname{}, the first training-free PTQ framework that uniformly quantizes both the language backbone and the DiT action head of a VLA model to W4A4, without mixed-precision allocation. \methodname{} combines a composite SVD$\cdot$Hadamard rotation that diffuses weight and activation outliers with a per-step DiT activation calibration that absorbs dynamic-range drift across denoising steps. On LIBERO, \methodname{} reaches 87.8\% on GR00T N1.5 and 98.0\% on Pi-0.5 under W4A4---matching or exceeding the FP16 references while shrinking the static footprint by 71.3\%---and real-world bimanual experiments confirm smoother, more reliable control than prior W4A4 methods. These results overturn the prevailing belief that the DiT action head is too sensitive to uniformly quantize.

\section*{Limitations}
Our evaluation focuses on two DiT-based VLA backbones (Pi-0.5 and GR00T N1.5) and one benchmark family (LIBERO + an ARX R5 real-world setup); broader validation on autoregressive or flow-matching action heads and on additional hardware platforms is left for future work. The vision encoder (ViT) is kept at FP16 in our current pipeline, since it accounts for a comparatively small fraction of the inference cost; extending the composite rotation to the vision encoder is a natural next step. Real-world results are averaged over 10 rollouts per task, which is sufficient to expose qualitative behavioral differences but leaves non-trivial statistical variance on the individual progress scores. We report static memory savings and end-to-end task performance, but not wall-clock latency: realizing the throughput benefits of W4A4 requires kernel-level support that is hardware- and toolchain-dependent and beyond the scope of this paper. Finally, we focus on the W4A4 regime as the current sweet spot between hardware support and accuracy; pushing further to W3 or W2 precision---where outlier suppression and per-step calibration alone are unlikely to suffice---is a promising direction that may require complementary techniques such as low-rank residual branches or learned codebooks.
\section*{AI Tool Usage}
During the preparation of this manuscript, the authors only used
AI tools for refining the language and presentation
of the written text. All scientific content, experimental
design, analysis, and conclusions were produced entirely by
the authors. The authors reviewed and edited all
AI-assisted output and take full responsibility for the
final manuscript.

\bibliography{custom, main}

\appendix

\section{Appendix}

\subsection{Description of Baselines and Benchmarks}
\label{sec:baselines}

\paragraph{Benchmark.}
We evaluate all methods on LIBERO~\citep{liu2023libero}, a standard benchmark for VLA policy evaluation in robotic manipulation. LIBERO consists of four task suites:
\begin{itemize}
    \item \textbf{Goal}: goal-conditioned manipulation tasks that require following language instructions.
    \item \textbf{Spatial}: tasks emphasizing spatial relations and precise object placement.
    \item \textbf{Object}: object-centric manipulation tasks involving object recognition and interaction.
    \item \textbf{Long}: long-horizon tasks requiring multiple sequential manipulation steps.
\end{itemize}
We report the success rate on each suite and the four-suite average.

\paragraph{Baselines.}
We compare \methodname{} against representative post-training quantization baselines:
\begin{itemize}
    \item \textbf{GPTQ}~\citep{frantar2022gptq}: a second-order, layer-wise weight quantization method that minimizes reconstruction error using calibration activations.
    \item \textbf{AWQ}~\citep{lin2024awq}: an activation-aware weight quantization method that protects salient channels based on activation statistics.
    \item \textbf{OmniQuant}~\citep{shao2023omniquant}: a PTQ method that optimizes clipping ranges and scaling factors through layer-wise reconstruction.
    \item \textbf{SmoothQuant}~\citep{xiao2023smoothquant}: a weight-activation quantization method that migrates activation outliers into weights before quantization.
    \item \textbf{DuQuant}~\citep{lin2024duquant}: a rotation-based PTQ method that improves low-bit weight-activation quantization through distribution-aware transformations.
    \item \textbf{QuantVLA}~\citep{quantvla}: a VLA-specific quantization baseline that combines low-bit quantization with VLA-oriented calibration components.
\end{itemize}
\paragraph{Calibration and evaluation protocol.}
For all quantization methods that require calibration data (including \methodname{} and all baselines), we use the \emph{same} unlabeled calibration buffer for fairness. The buffer consists of 10 trajectories per task suite---one trajectory from each of the 10 tasks within the suite, sampled from a single fixed initial state---collected by rolling out the FP16 base model in simulation. We then evaluate on 10 trials per task with initial states held out from calibration, ensuring no overlap between calibration and evaluation conditions. The success rates reported in Table~\ref{tab:ns10_quant_results} are averaged across all tasks within each suite.
\subsubsection{Key Hyperparameters}
\label{sec:key_hyperparameters}
Table~\ref{tab:key_method_configs} summarizes the configurations of all methods reported in the main LIBERO comparison (Table~\ref{tab:ns10_quant_results}).

\begin{table*}[t]
\centering
\small
\setlength{\tabcolsep}{4pt}
\renewcommand{\arraystretch}{1.18}
\caption{Key configurations of all methods reported in the main LIBERO comparison.}
\label{tab:key_method_configs}
\vspace{0.4em}
\resizebox{\textwidth}{!}{%
\begin{tabular}{llp{0.64\linewidth}}
\toprule
\textbf{Method} & \textbf{Setting} & \textbf{Key hyperparameters / runtime configuration} \\
\midrule

FP16 & FP16
& Original full-precision model without quantization; used as the reference upper bound. \\

\midrule
GPTQ~\citep{frantar2022gptq} & W4A16
& Weight-only quantization with FP16 activations. Block size $128$, damping ratio $0.01$, Hessian-aware layer-wise reconstruction. The input Hessian is estimated from the calibration activations. \\

AWQ~\citep{lin2024awq} & W4A16
& Weight-only activation-aware quantization. Activation-based channel saliency, grid-based scale search with $20$ candidates, maximum shrink ratio $0.5$, and $512$ sampled calibration tokens for clipping and scale search. \\

OmniQuant~\citep{shao2023omniquant} & W4A16
& Weight-only PTQ with optimized clipping and scaling parameters. Layer-wise reconstruction with learnable quantization parameters, optimized with Adam at learning rate $5{\times}10^{-3}$ for $200$ iterations. \\

\midrule
SmoothQuant~\citep{xiao2023smoothquant} & W4A8 / W4A4
& Weight-activation quantization with smoothing coefficient $\alpha=0.5$ and clipping threshold $10^{3}$. The per-channel smoothing scale is estimated from calibration activations and absorbed into the weight matrix before quantization. Activations are quantized to 8 or 4 bits according to the setting. \\

DuQuant~\citep{lin2024duquant} & W4A8 / W4A4
& Rotation-based weight-activation quantization. Block-wise rotation with block size $64$, SVD-based rotation, weight-energy channel permutation, and row-rotation restore. Activation calibration uses percentile $p=99.9$ over calibration steps. \\

QuantVLA~\citep{quantvla} & W4A4
& VLA-specific quantization baseline built on DuQuant-style low-bit quantization, with DuQuant defaults, ATM calibration, and Output-Head Boosting as proposed in the original paper. Following the released setting, DiT attention layers are kept in FP16 unless otherwise noted. \\

\midrule
\textbf{\methodname{}} & \textbf{W4A4}
& Uniform W4A4 quantization across the LLM backbone and DiT action head. \textit{Rotation:} block-wise SVD$\cdot$Hadamard rotation with block size $64$, preceded by a zigzag weight-norm channel permutation (Sec.~\ref{sec:method}). \textit{Solver:} GPTQ-style weight quantization on the LLM side (block size $128$, damping ratio $0.01$); plain RTN on the DiT side, motivated by the layer-wise error asymmetry analysis in Appendix~\ref{sec:solver_selection}. \textit{Activation calibration:} per-step, per-channel scale table calibrated over $T=8$ Euler denoising steps on $n=10$ unlabeled calibration trajectories. All scales are folded into dequantization at inference. \\

\bottomrule
\end{tabular}
}
\end{table*}

\subsection{Real-World Manipulation Tasks}
\label{sec:real_tasks}
To evaluate \methodname{} in real-world scenarios, we construct five manipulation tasks with varying levels of difficulty, ranging from simple object placement to long-horizon bimanual manipulation. As summarized in Table~\ref{tab:real_tasks}, each task is decomposed into multiple sequential stages and evaluated using a progress score, enabling a fine-grained assessment of task completion. Detailed scoring criteria are provided in Table~\ref{tab:real_tasks}.
\begin{figure*}[!t]
  \centering
  \includegraphics[width=0.98\textwidth]{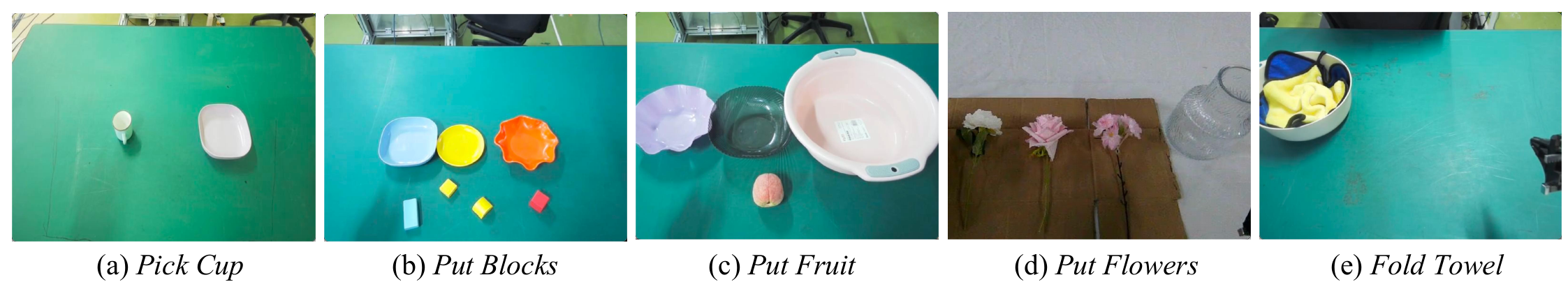}
  \caption{
  Experimental setup of the real-world manipulation tasks.
  (a) \textit{Pick Cup}: move the plate to the center and place the cup onto it.
  (b) \textit{Put Blocks}: place colored blocks into their matching plates.
  (c) \textit{Put Fruit}: place the fruit into the designated container according to the language instruction.
  (d) \textit{Put Flowers}: insert three flowers into a vase.
  (e) \textit{Fold Towel}: pick up, flatten, fold, and place the towel back.
  }
  \label{fig:real-task-setting}
\end{figure*}
\begin{table*}[t]
\centering
\small
\renewcommand{\arraystretch}{1.4}
\begin{tabular}{>{\centering\arraybackslash}p{0.8cm}
                >{\centering\arraybackslash}p{2.3cm}
                p{5.8cm}
                p{6.5cm}}
\toprule
\textbf{Task ID} &
\textbf{Task Name} &
\textbf{Task Description} &
\textbf{Scoring Criteria} \\
\midrule

1 &
Pick Cup &
Place a cup onto a plate using bimanual coordination. &
40 points are awarded for grasping the cup with the right arm, 30 points for moving the plate to the center with the left arm, 20 points for placing the cup onto the plate, and 10 points for returning both arms to the home position. \\
\midrule

2 &
Put Blocks &
Place four colored blocks into the plates with matching colors. &
25 points are awarded for each correctly matched block, resulting in a maximum score of 100 points. \\
\midrule

3 &
Put Fruit &
Place the target fruit into the corresponding container according to the language instruction. &
40 points are awarded for grasping the target fruit, 50 points for placing it into the correct container, and 10 points for returning both arms to the home position. \\
\midrule

4 &
Put Flowers &
Insert three flowers into a vase. &
For each flower, 15 points are awarded for grasping and 15 points for successful insertion. Completing all three flowers yields 90 points, while returning both arms to the home position yields an additional 10 points. \\
\midrule

5 &
Fold Towel &
Fold a towel and place it back to the designated area. &
20 points are awarded for picking up the towel, 30 points for flattening it, 30 points for completing the folding operation, and 20 points for placing the folded towel back in the designated area. \\
\bottomrule
\end{tabular}
\caption{Descriptions and progress-score criteria for the five real-world manipulation tasks. Each task is decomposed into multiple stages, and the progress score is computed according to the completion of intermediate subtasks.}
\label{tab:real_tasks}
\end{table*}

\subsection{Open-Loop Action Analysis}
\label{sec:action_analysis}
Figure~\ref{fig:action_vis} presents an open-loop comparison of the predicted action trajectories on real-world tasks. The blue dashed curves denote the proposed \methodname{}, while the red curves correspond to the baseline quantization method, QuantVLA. Specifically, we feed observations from the dataset into the model and visualize the predicted actions across all 14 action dimensions. Although both quantized models introduce prediction deviations relative to the ground truth, \methodname{} consistently generates action trajectories that more closely match the reference actions. As shown in the figure, the baseline method occasionally produces noticeable spikes and abrupt deviations in several action dimensions, indicating that quantization errors may be amplified into extreme action outputs. In contrast, \methodname{} yields smoother and more consistent trajectories, suggesting that the proposed quantization strategy better preserves the original action distribution. Similar trends can also be observed in the gripper action dimensions, where \methodname{} exhibits smaller fluctuations and more stable predictions. Such improvements are particularly important for closed-loop manipulation, as unstable action outputs may accumulate over time and result in jerky motions or inaccurate object interactions. We believe this observation partially explains why \methodname{} achieves superior real-world manipulation performance under the same W4A4 quantization setting.

\subsection{Selection of Weight Quantization Solvers} 

\label{sec:solver_selection}
\begin{figure*}[!t]
  \centering
  \includegraphics[width=0.98\textwidth]{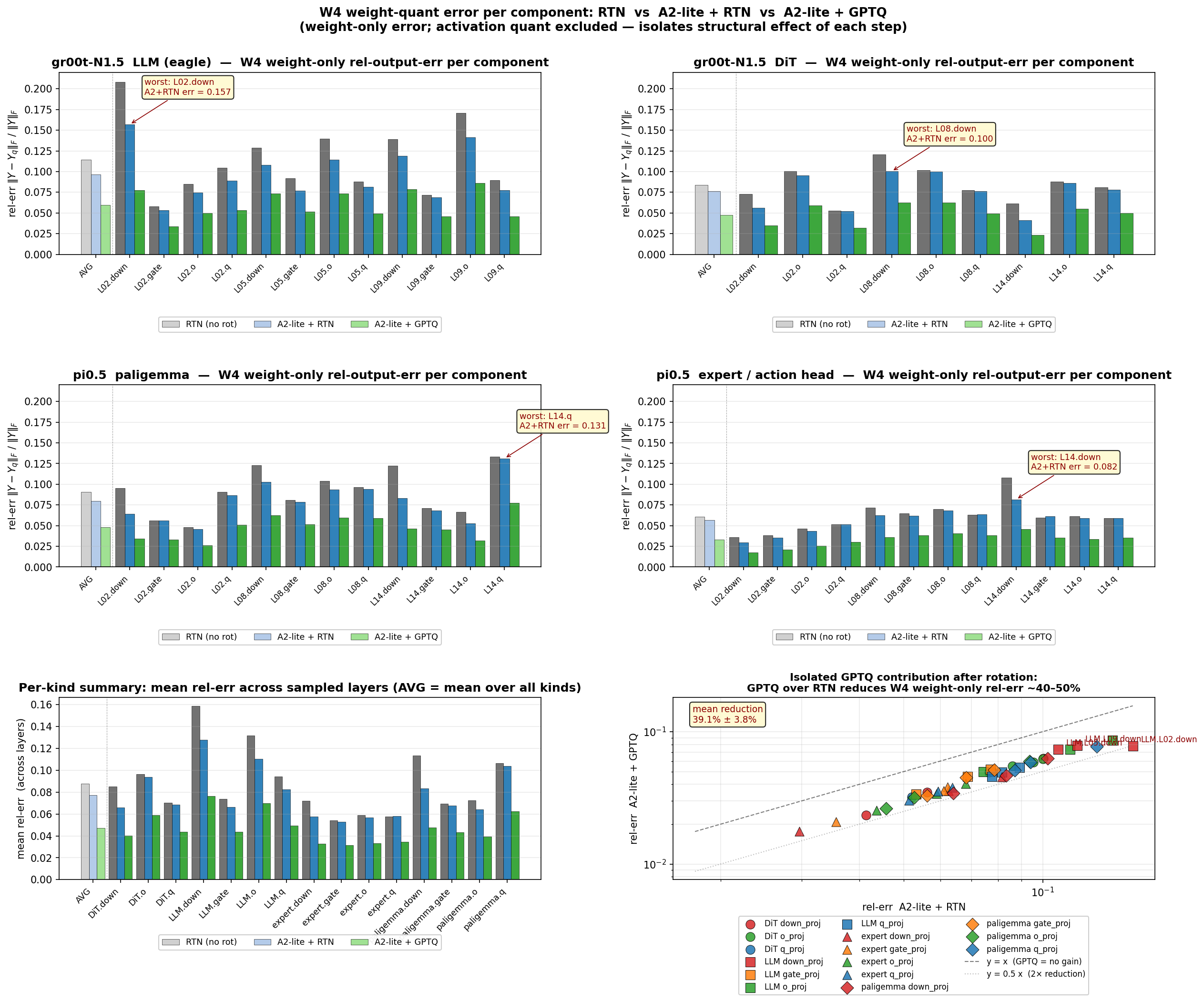}
  \caption{
  Per-component W4 weight quantization error across gr00t-N1.5 and pi0.5. Relative output error $|XW^\top - XR,Q(WR)^\top|_F / |XW^\top|_F$ measured layer-by-layer for three weight pipelines: RTN with no rotation, SVD-Hadamard rotation + RTN, and SVD-Hadamard rotation + GPTQ. Activation quantization is excluded so the bars isolate the structural contribution of each weight-quant step.
  }
  \label{fig:quant_err_per_component}
\end{figure*}

Figure~\ref{fig:quant_err_per_component} measures the layer-wise W4 weight-only output error after the SVD-Hadamard rotation, with and without an additional GPTQ pass on the residual. Two observations stand out.

First, GPTQ unambiguously reduces the numerical per-layer quantization error on both sides of both models. Across all 16 module-kind combinations on gr00t and pi0.5, SVD-Hadamard + GPTQ has a strictly lower bar than SVD-Hadamard + RTN — roughly a 40-50\% reduction in relative output error on average, consistent with what GPTQ-style inter-column error compensation is supposed to deliver after activation outliers have been spread by rotation.

Second, the absolute quantization difficulty after rotation is asymmetric between the two sides. On the LLM side (gr00t eagle, pi0.5 paligemma), even after SVD-Hadamard the relative error remains in the 0.07-0.13 range, with the LLM down\_proj layers consistently the hardest — gr00t LLM.L02/L05/L09.down\_proj all sit above 0.10, reflecting the residual outlier structure that survives Hadamard mixing. On the DiT/expert side (gr00t action head, pi0.5 expert), SVD-Hadamard alone already brings most layers to 0.04-0.07; GPTQ pushes them further down, but the absolute headroom GPTQ can claim is much smaller than on the LLM side.

This per-side asymmetry maps directly onto what we observe end-to-end: adding GPTQ on the LLM side yields a measurable LIBERO success gain, because rotation alone does not eliminate the LLM down\_proj difficulty and GPTQ's column-wise compensation buys real precision back. On the DiT side, however, the same numerical improvement does not translate into a downstream score gain; we attribute this to the fact that, with the SVD-Hadamard rotation already producing near-uniform per-row column magnitudes on the DiT weights, GPTQ's inter-column error propagation has little structure left to exploit and instead introduces a calibration-dependent bias that interacts poorly with the per-step diffusion dynamics. The figure therefore makes concrete the asymmetric design choice underlying \methodname{}: GPTQ is applied only on the LLM side, while the DiT side retains plain RTN.

\subsection{Effectiveness of the SVD-Hadamard Rotation} 
\label{app:hadamard}
\begin{figure*}[!t]
  \centering
  \includegraphics[width=0.98\textwidth]{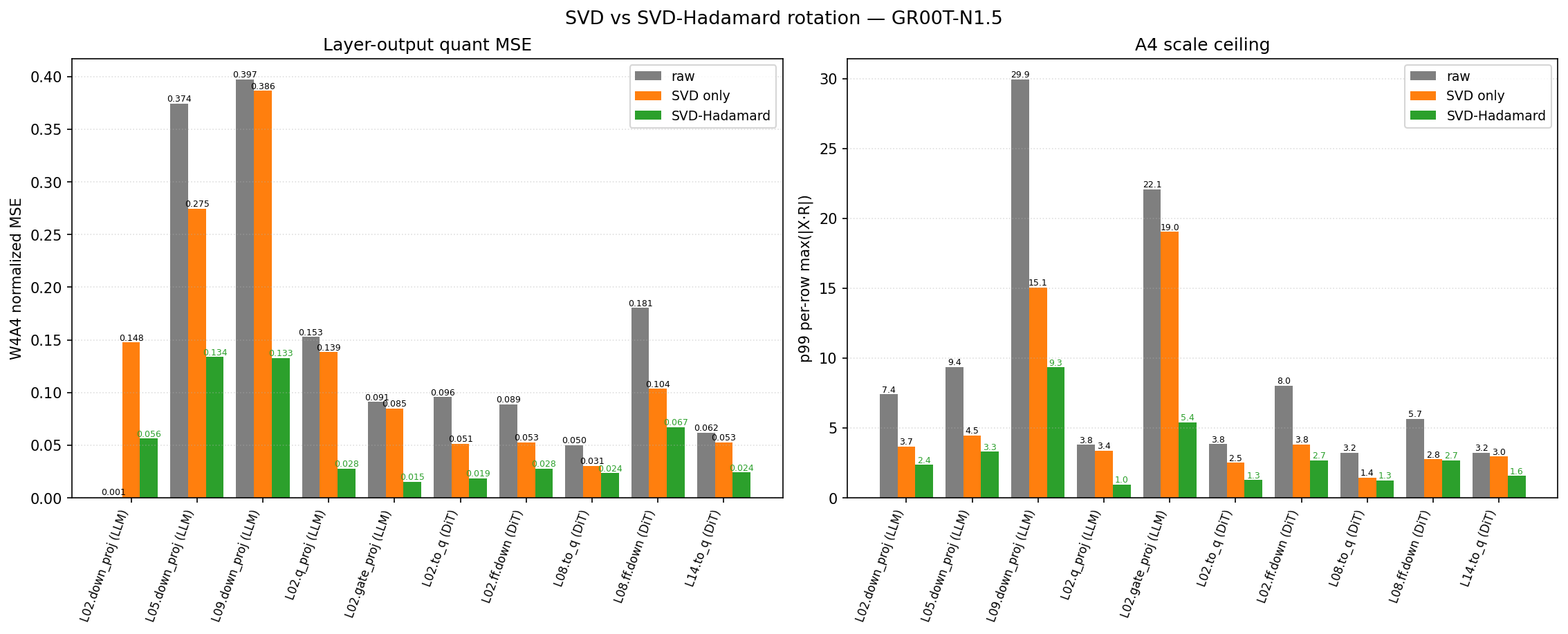}
  \caption{
  Per-layer W4A4 quantization quality on GR00T-N1.5. Left: normalized MSE between FP16 and W4A4 layer outputs. Right: 99th-percentile per-row max of rotated activations (sets the A4 scale ceiling). Bars compare raw, SVD-only, and SVD-Hadamard (A2-lite) rotation across 5 LLM + 5 DiT sampled layers. SVD-Hadamard yields the lowest nMSE and lowest scale ceiling on 9/10 layers. The single exception is LLM.L02.down\_proj, whose extreme channel-skew $(\approx 50000×)$ aligns with the SVD basis and is only partially mitigated by the within-block Hadamard mixing.}
  \label{fig:svd_vs_svdh}
\end{figure*}

Figure~\ref{fig:svd_vs_svdh} quantifies the gain from adding the Hadamard tail to the per-block SVD rotation, measured on 10 sampled GR00T layers (5 LLM + 5 DiT) under a true W4A4 quantization sweep. Two complementary metrics are reported: the left panel shows the end-to-end normalized MSE between the FP16 and the W4A4 layer output (what the downstream forward pass actually consumes), while the right panel shows the 99th-percentile per-row maximum of the rotated activation $X \cdot R$, which sets the A4 scale on the heaviest rows and therefore upper-bounds how much dynamic range A4 must accommodate. The two metrics agree closely.

For 9 of the 10 sampled layers, SVD-Hadamard yields both the lowest layer-output nMSE and the lowest A4 scale ceiling, with the gain over SVD-only often exceeding 2-5$\times$ — for example, LLM.L02.q\_proj drops from $0.139$ (SVD-only) to $0.028$ (SVD-Hadamard), and DiT.L02.q\_proj from $0.051$ to $0.019$. This confirms the geometric intuition that an SVD basis fitted block-by-block on the weight matrix does not, by itself, address activation-side outliers; the Hadamard mixing applied on top spreads any residual single-channel concentration across the 64 channels of the block, simultaneously lowering both the per-row dynamic range and the resulting W4A4 reconstruction error.

\begin{figure*}[!t]
  \centering
  \includegraphics[width=0.98\textwidth]{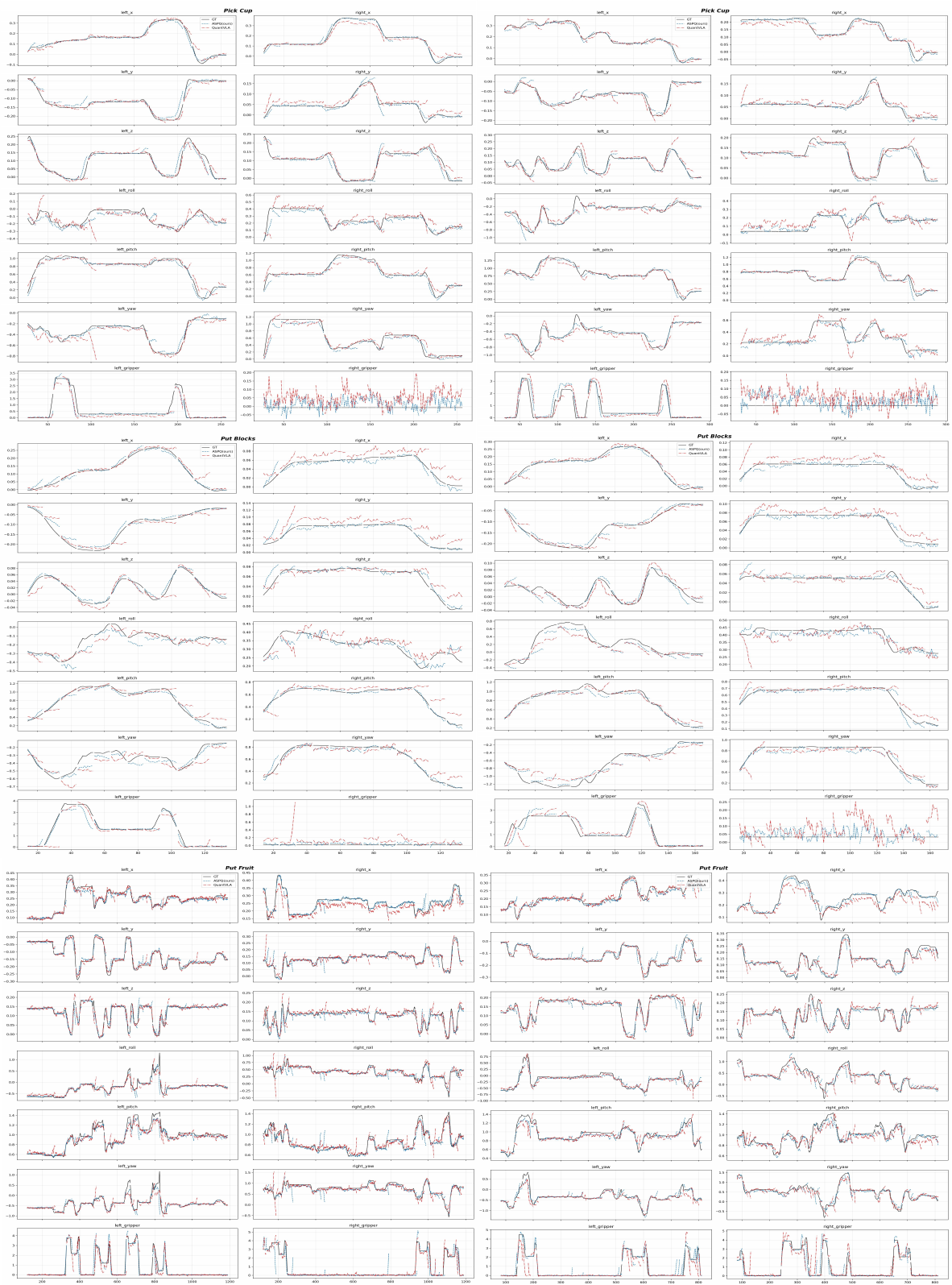}
  \caption{
  Open-loop action trajectory comparison across all 14 action dimensions. The blue dashed curves denote \methodname{} and the red curves denote QuantVLA. \methodname{} generates trajectories that more closely match the ground-truth trajectories and exhibits fewer spikes, abrupt deviations, and extreme outputs. Such improved action stability may reduce control jitter during closed-loop execution, providing a plausible explanation for the superior real-world manipulation performance achieved by \methodname{}.
  }
  \label{fig:action_vis}
\end{figure*}

The one revealing exception is LLM.L02.down\_proj, a layer with channel-skew on the order of $5 \times 10^{4}$. Here pure SVD is actively harmful: its nMSE inflates from $0.001$ on raw to $0.148$ — roughly two orders of magnitude worse — because the dominant SVD direction aligns with the outlier channel and concentrates rather than disperses it. SVD-Hadamard restores partial sanity ($0.056$) but does not fully recover the raw baseline, indicating that for this extreme outlier the Hadamard $\sqrt{64} \approx 8\times$ attenuation is not enough on its own. The picture here is consistent with the per-(token,block) heatmap evidence (Figure~\ref{fig:action_vis}) and explains why our final pipeline pairs the SVD-Hadamard rotation with calibrated W-side GPTQ on the LLM, while the DiT side — where rotation alone already pushes nMSE below the level at which inter-column compensation could help — uses plain RTN.

\subsection{Necessity of Per-Step Activation Scaling for DiT} 
\label{app:per_step_scale}
\begin{figure*}[t]
  \includegraphics[width=1\linewidth]{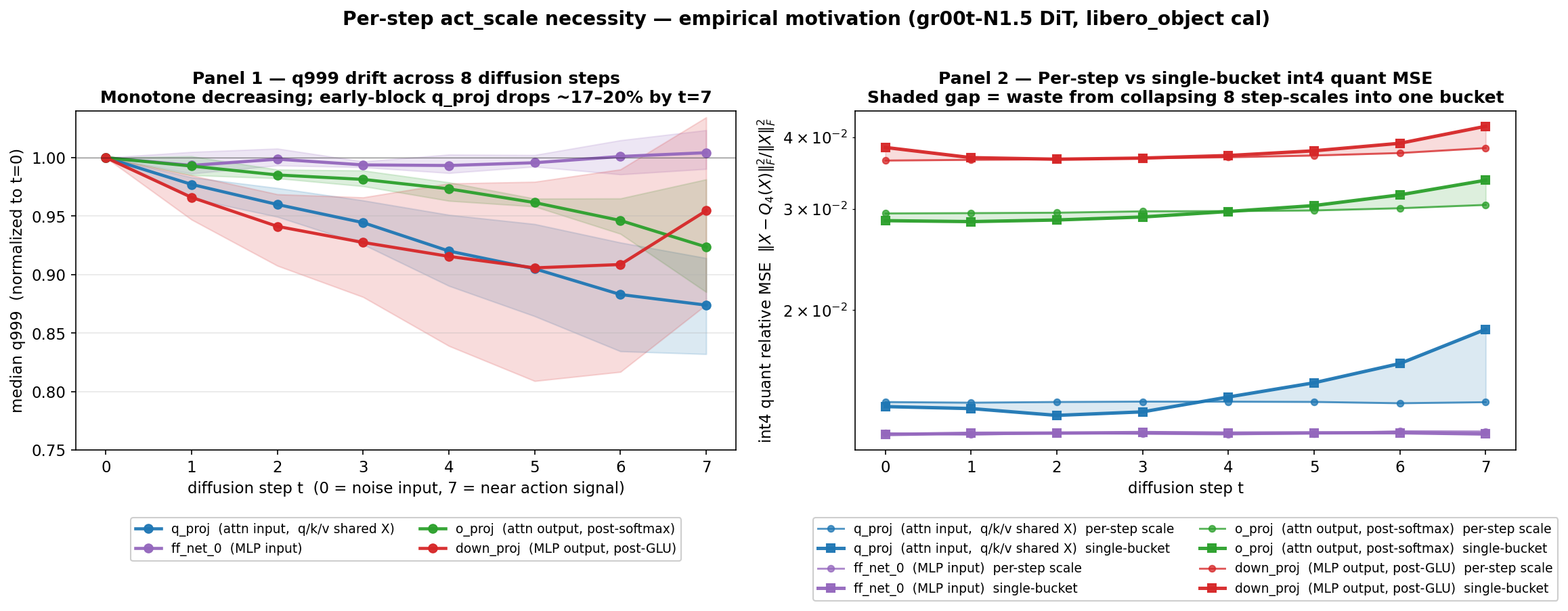}
              \caption{Necessity of per-step act\_scale on the DiT side. Left: per-channel q999 across the 8 Euler denoising steps for four DiT layer kinds (normalized to $t=0$). Right: int4 quantization MSE per step under a step-specific scale (circles) vs a single-bucket scale (squares); shaded area = waste from collapsing 8 step-scales into one. Time-conditioned magnitude drift only appears at layers reading the \texttt{AdaLayerNorm} output (attention QKV projections), and per-step scaling helps exactly those layers.}

    \label{fig:perstep_necessity}
\end{figure*}

Figure~\ref{fig:perstep_necessity} (left) plots the 99.9-percentile activation magnitude across the 8 Euler denoising steps for four sampled DiT layer kinds. We observe a sharp layer-type asymmetry: \texttt{attn1.to\_q/k/v} (the QKV projections that follow \texttt{norm1}) exhibit a monotonic 15--20\% drop in q999 from step $t=0$ (pure noise input) to step $t=7$ (near-converged action signal), while \texttt{ff.net[0]} (the MLP projection following \texttt{norm3}) is essentially flat (within 2\%). The remaining attention output (\texttt{attn1.to\_out}) and MLP output (\texttt{ff.net[2]}, \texttt{down\_proj}) sit in between with 5--10\% drift. The source-code structure of the DiT block explains this asymmetry: \texttt{norm1} is an \texttt{AdaLayerNorm} whose per-step scale and shift are functions of the timestep embedding $t$, so its output explicitly carries the time-conditioned magnitude into the QKV input; \texttt{norm3} is a plain \texttt{LayerNorm} that strips time-dependent magnitude variation away before the MLP. The layers that benefit from per-step act\_scale are exactly those that read a time-conditioned activation.

The right panel translates this distributional drift into a quantization cost. For each DiT layer $\times$ diffusion step, we measure the relative MSE of int4 symmetric quantization under (i) a step-specific per-channel scale (computed from that step's own q999) and (ii) a single-bucket scale (the mean q999 over all 8 steps). Per-step MSE stays low and flat across the 8 steps; single-bucket MSE rises at the steps whose q999 deviates most from the across-step mean, with the shaded gap quantifying the wasted int4 budget. Aggregating across the 4 layer kinds $\times$ 3 depth samples $\times$ 8 steps gives a mean per-(layer, step) MSE gap of \textbf{2.5\%}, dominated by \texttt{q\_proj}-style layers (gap up to 11\%) and essentially zero for \texttt{ff.net[0]}. Compounded over the $\sim$16 DiT blocks and 8 Euler steps of a single action prediction, this per-step waste maps empirically to an end-to-end loss of \textbf{$-2.0$ pp 4-suite average and $-7.0$ pp on the long-horizon suite} (with vs.\ without per-step scaling; rows 2 and 3 of Table~\ref{tab:ablation_combined}), confirming that per-step \texttt{act\_scale} is necessary in the W4A4 regime---and confirming that the necessity is concentrated on the post-\texttt{AdaLayerNorm} attention inputs rather than uniformly across all DiT linears.

\begin{figure*}[h!]
    \centering
    \includegraphics[width=0.85\linewidth]{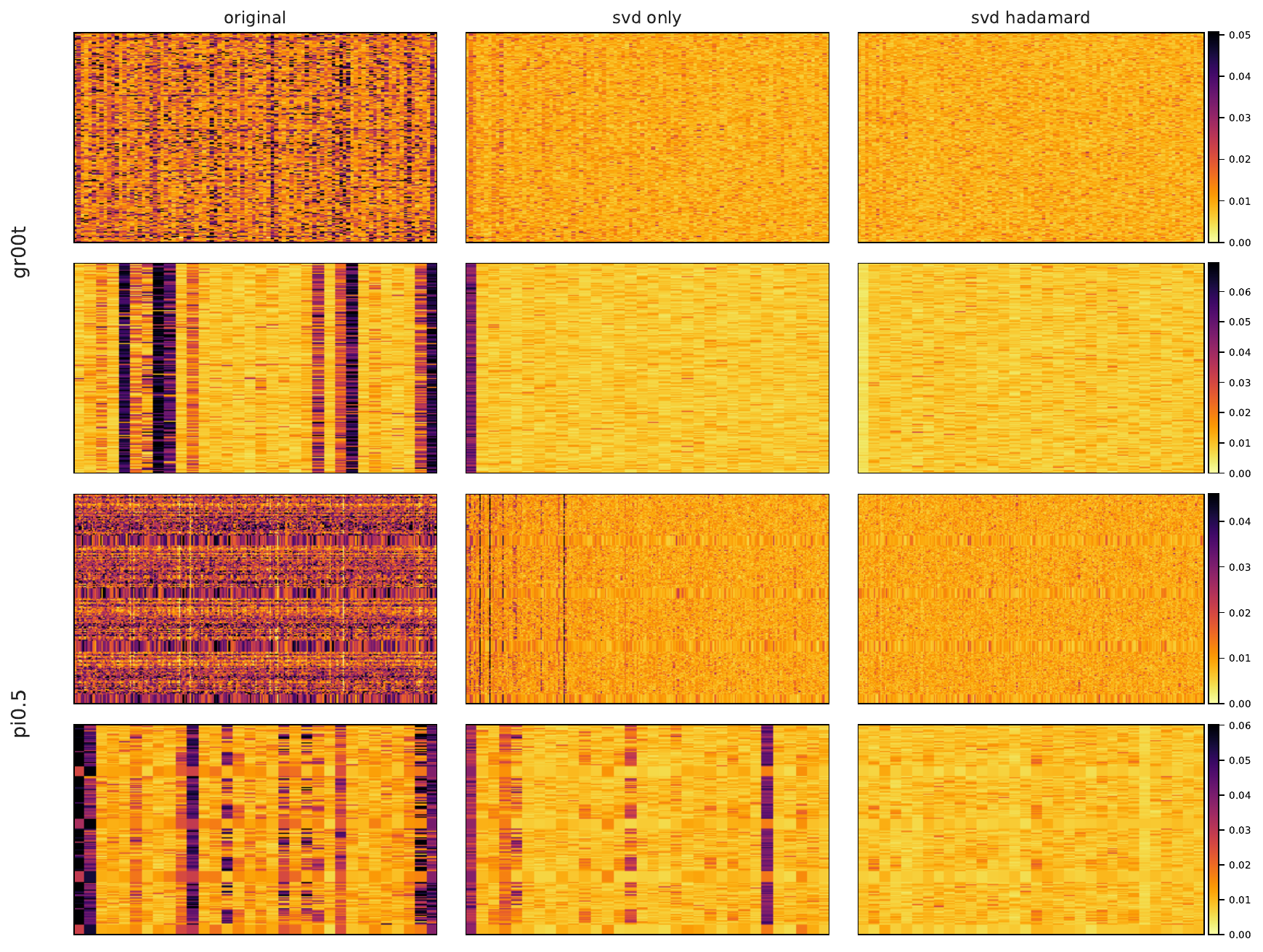}
    \caption{The quantization error with different Rotation methods.}
    \label{fig:activation_outlier}
\end{figure*}
\section{Quantization Error with Different Rotation}
\label{app:rotation_error}

Figure~\ref{fig:activation_outlier} further compares the layer-wise quantization
error under different rotation strategies on both GR00T-N1.5 and Pi-0.5. We
evaluate three settings: the original unrotated representation, SVD-only
rotation, and the proposed SVD-Hadamard rotation. The error is measured after
low-bit quantization and reflects how well each transformed representation
preserves the full-precision layer output.

Across both backbones, SVD-Hadamard generally yields lower quantization error
than the unrotated baseline and the SVD-only variant. This trend supports the
motivation of our composite rotation: SVD reduces weight-side channel imbalance,
while the Hadamard transform further diffuses residual activation outliers that
are not necessarily aligned with the SVD basis. In contrast, SVD-only rotation
can improve weight smoothness but does not consistently control activation-side
dynamic range, which explains its less stable behavior across layers and models.

\subsection{Real-Quantization Kernel Implementation}

For real low-bit inference, we implement a W4A8 execution backend tailored to our rotated weights. The learned rotations are folded into the linear weights offline, after which the weights are quantized to INT4 and converted to the packed layout required by the CUDA GEMM kernel. At runtime, FP16 activations are dynamically quantized to INT8 with per-token scales and multiplied with the packed INT4 weights using an INT8 Tensor-Core GEMM. We expose this path as a compile-compatible PyTorch custom operator, with per-channel scaling and zero-point correction handled in the kernel interface. Since the rotations are absorbed during offline weight preparation, no additional rotation operation is introduced at runtime.
\begin{table}[t]
\centering
\caption{Real-quantization inference efficiency on an NVIDIA H100 GPU. End-to-end (E2E) latency is measured on Spatial Task 0 and averaged over three episodes under the same serving setup.}
\label{tab:real_quant_latency}
\resizebox{\columnwidth}{!}{
\begin{tabular}{lcc}
\toprule
\textbf{Method} & \textbf{Per-step (ms)} & \textbf{E2E / Episode (s)} \\
\midrule
FP16 & 32.2 & \textbf{8.8} \\
$\Omega$-QVLA (Ours) & \textbf{26.7} & \textbf{12.4} \\
QVLA-real & 27.1 & 19.9 \\
QuantVLA-real & 29.8 & 19.9 \\
\bottomrule
\end{tabular}
}
\end{table}

Our real-quantized implementation achieves the fastest per-step inference latency (\textbf{26.7 ms}) among all settings, including FP16, and is also substantially faster than competing real-quantized methods in end-to-end evaluation. The remaining gap between our end-to-end latency and FP16 is attributed to serving-side overhead shared by the real-quantization backend rather than the low-bit GEMM itself.

\end{document}